\documentclass[conference]{IEEEtran}
\IEEEoverridecommandlockouts
\usepackage{caption}
\usepackage{multirow}
\usepackage{xcolor}
\usepackage{graphicx}
\usepackage{epstopdf}
\usepackage{graphics}
\usepackage{url}
\usepackage{cuted}     
\usepackage{caption}
\usepackage{cite}
\usepackage{comment}
\usepackage{stfloats}
\usepackage{float} 
\usepackage{booktabs}
\usepackage{amsmath}
\DeclareUnicodeCharacter{200B}{}
\AtBeginDocument{%
  \providecommand\BibTeX{{%
    \normalfont B\kern-0.5em{\scshape i\kern-0.25em b}\kern-0.8em\TeX}}
    }
\makeatletter 
\newcommand{\linebreakand}{%
  \end{@IEEEauthorhalign}
  \hfill\mbox{}\par
  \mbox{}\hfill\begin{@IEEEauthorhalign}
}
\makeatother 

\begin{document}

\title{Memoria: A Scalable Agentic Memory Framework for Personalized Conversational AI}

\author{\IEEEauthorblockN{Samarth Sarin}
\IEEEauthorblockA{\textit{BlackRock, Inc.} \\
Gurgaon, HR, India \\
samarth.sarin@blackrock.com}
\and
\IEEEauthorblockN{Lovepreet Singh}
\IEEEauthorblockA{\textit{BlackRock, Inc.} \\
Gurgaon, HR, India \\
lovepreet.singh@blackrock.com}
\and 
\IEEEauthorblockN{Bhaskarjit Sarmah}
\IEEEauthorblockA{\textit{BlackRock, Inc.} \\
Gurgaon, HR, India \\
bhaskarjit.sarmah@blackrock.com}
\and 
\linebreakand 
\IEEEauthorblockN{Dhagash Mehta}
\IEEEauthorblockA{\textit{BlackRock, Inc.} \\
New York, NY, USA \\
dhagash.mehta@blackrock.com}
}

\maketitle

\begin{abstract}
  Agentic memory is emerging as a key enabler for large language models (LLM) to maintain continuity, personalization, and long-term context in extended user interactions, critical capabilities for deploying LLMs as truly interactive and adaptive agents. Agentic memory refers to the memory that provides an LLM with agent-like persistence: the ability to retain and act upon information across conversations, similar to how a human would. We present Memoria, a modular memory framework that augments LLM-based conversational systems with persistent, interpretable, and context-rich memory. Memoria integrates two complementary components: dynamic session-level summarization and a weighted knowledge graph (KG)-based user modelling engine that incrementally captures user traits, preferences, and behavioral patterns as structured entities and relationships. This hybrid architecture enables both short-term dialogue coherence and long-term personalization while operating within the token constraints of modern LLMs. We demonstrate how Memoria enables scalable, personalized conversational artificial intelligence (AI) by bridging the gap between stateless LLM interfaces and agentic memory systems, offering a practical solution for industry applications requiring adaptive and evolving user experiences.
\end{abstract}


\maketitle

\section{Introduction}
Large Language Models (LLMs)\cite{chang2024survey} have significantly advanced the capabilities of conversational AI, enabling human-like interactions across a broad spectrum of applications. However, most LLM-based chat systems operate without persistent memory, where each interaction is treated in isolation, discarding previous context and failing to adapt over time. This limits their ability to form coherent, context-rich, and personalized conversations.

Incorporating persistent memory into LLM-powered systems offers substantial business and operational value by enhancing user experience, reducing interaction friction, and enabling context-aware automation. In customer support domains, memory-equipped agents can retrieve prior user interactions, historical issues, and stated preferences, thereby improving resolution speed, reducing redundant information gathering, and fostering a more personalized dialogue experience. In domains like customer support, e-commerce, and financial advisory, memory allows faster resolutions, customized recommendations, and long-term personalization.

Agentic memory\cite{zhang2024survey} refers to the capacity of an LLM agent implementation to recall, adapt to, and reason over past interactions to simulate the behaviour of a coherent, goal-directed agent. It allows LLMs to move beyond reactive, stateless responses toward sustained, context-aware dialogues. Incorporating such memory into LLM-based systems is not only crucial for technical performance but also vital for enhancing user trust, reducing repetition, and improving long-term engagement.

The development of agentic memory in LLM-based systems has become a key research focus. Zhang et al.\cite{zhang2024survey} provide a comprehensive survey categorizing memory types (textual, parametric, structured) and operations like writing, retrieval, and forgetting, while calling for unified frameworks. A‑MEM\cite{xu2025mem}, inspired by Zettelkasten note-taking~\cite{luhmann1992communicating}, introduces graph-like memory structures that link atomic notes with contextual descriptors, enabling dynamic memory evolution beyond flat storage systems.

On the systems side, knowledge-graph (KG)-based memory architectures have gained traction in enabling persistent, interpretable memory for conversational agents. Batching user interactions into temporal KGs, as done in Ref.~\cite{rasmussen2025zep}, demonstrates a practical pathway for capturing evolving user states. Similarly, industrial frameworks such as LangGraph\cite{langgraph2025} and LlamaIndex\cite{llamaindex2024} have begun adopting summarization, vector retrieval, and graph-based profiling. 

Existing LLM memory systems address components like vector retrieval, summarization, or temporal KGs in isolation. Vector stores lack interpretability and conflict resolution, while graph-based systems struggle with recency and scalability. Few frameworks integrate both short- and long-term memory with incremental, recency-aware updates.

To fill this gap, we propose Memoria, a (Python-based) memory augmentation framework that can be integrated into any LLM-driven chat interface. Memoria introduces a persistent memory layer through two complementary components: (1) dynamic chat summarization and (2) a weighted KG-based user persona engine. By combining these components, Memoria enables context retention and behavioral continuity across sessions.

\section{Background and Proposed Methodology}
\subsection{Background and Related Work}
Recent research has increasingly recognized agentic memory as a critical capability for enhancing LLM-based conversational systems. Traditional LLM deployments typically follow a stateless architecture in which each user input is processed independently, with prior interactions forgotten unless explicitly provided as input context. This design leads to repetitive, impersonal exchanges that fail to leverage historical user information. Agentic memory frameworks seek to overcome these limitations by equipping models with mechanisms to retain, retrieve, and reason over accumulated conversational histories, enabling long-term personalization and adaptive dialogue behavior. By maintaining structured memory representations such as KGs, session summaries, or retrieved interaction histories, agentic systems can evolve beyond reactive responses to support proactive, goal-aligned interactions that more closely resemble long-term human agency\cite{zhang2024survey, xu2025mem, rasmussen2025zep}.

Below are various types of agentic memories crucial for a personalized conversational AI system:

\noindent\textbf{Episodic Memory:} Episodic memory\cite{tulving1972episodic,zhang2024survey} encapsulates the agent’s ability to recall specific past interactions or events. This type of memory mirrors the autobiographical memory of humans, allowing the model to reference details such as user preferences, prior conversations, and historical decisions. For example, if a user is previously identified as an equity trader, an agent equipped with episodic memory could incorporate this information to personalize future financial updates. Such capabilities support persistent personalization and conversational continuity across sessions, both of which are critical for enhancing user engagement and interaction quality in long-term deployments.

\noindent\textbf{Semantic Memory:} Semantic memory\cite{tulving1972episodic,zhang2024survey} represents the model’s capacity to access and utilize structured, factual knowledge, including historical data, general world knowledge, domain-specific definitions, and taxonomic relationships. This type of memory allows LLM-based agents to ground their responses in verifiable information by integrating external knowledge sources such as knowledge bases, ontologies, or APIs. For instance, retrieving historical stock prices, referencing regulatory definitions, or explaining financial instruments exemplifies the application of semantic memory in financial advisory contexts. By providing access to consistent and externally validated information, semantic memory plays a foundational role in ensuring response accuracy, factual consistency, and domain correctness, particularly in high-stakes professional environments.

\noindent\textbf{Working Memory:} Working memory\cite{baddeley2020working,zhang2024survey} refers to the short-term memory capacity that supports ongoing reasoning, multi-step problem solving, and temporary information retention during active tasks. This transient memory enables LLM-based agents to maintain intermediate computations, dialogue states, and contextual dependencies across multiple conversational turns. Analogous to human cognitive working memory, it allows the model to mentally track evolving information while formulating responses. Tasks such as iterative code debugging, multi-step mathematical reasoning, dynamic planning, or conversational task decomposition all benefit from robust working memory mechanisms within LLM architectures.

\noindent\textbf{Parametric Memory}
Parametric Memory~\cite{zhang2024survey} refers to the knowledge encoded in the parameters of a language model during pre-training. It enables zero- and few-shot generalization by capturing patterns and facts from the training data. While useful for tasks like explaining financial concepts without external context, it is static and cannot adapt to new user-specific interactions, limiting personalization.

These memory types form the foundation for agentic LLM systems. Recent work on external memory such as Retrieval Augmented Generation (RAG), summarization, and memory graphs aims to bridge short- and long-term context. Memoria builds on this by providing a developer-friendly framework that combines chat summarization, persistent logging, and KG-based user modeling.

\subsection{Proposed Methodology}
To overcome the limitations inherent in stateless LLM chat systems, particularly their inability to retain memory across sessions, personalize responses, or maintain coherent conversations over time, we propose \textbf{Memoria}: a modular Python-based library engineered to furnish LLMs with structured and persistent memory.

Memoria operates as an enhancement layer, integrating seamlessly into any LLM-driven conversational assistant architecture. Its design revolves around four core modules, each addressing a specific facet of memory and context awareness: (1) structured conversation logging, (2) dynamic user modeling, (3) real-time session summarization, and (4) context-aware retrieval. Together, these components form a unified memory framework that allows LLMs to simulate human-like memory retention and personalization. We propose solving the episodic and semantic memory capabilities of LLM applications. Our approach does not involve finetuning of an LLM, hence there is no enhancement provided in the parametric memory, and neither is there any enhancement provided by Memoria on working memory.

\subsubsection{Structured Conversation History with Database}

The foundational layer of Memoria is a \textbf{structured conversation logging system} powered by a database. Every user interaction is persistently stored in this database with a well-defined schema that includes:

\begin{itemize}
    \item \textbf{Timestamps} marking the exact time of each message;
    \item \textbf{Session identifiers} for differentiating between conversation instances;
    \item \textbf{Raw message content} from both the user and the LLM;
    \item \textbf{KG triplets} extracted from the user messages.
    \item \textbf{Summaries} of conversational turns for downstream retrieval and compression; and,
    \item \textbf{Token usage statistics} for monitoring efficiency and performance.
\end{itemize}
This structured format transforms the interaction history into a queryable and temporally indexed memory bank. By maintaining a comprehensive log of all interactions, Memoria enables the LLM to recall past user inputs even across different sessions. This persistent memory eliminates the need for users to reintroduce themselves or repeat previous inputs, thereby improving the continuity and flow of interaction. A key benefit of structured conversation is that users no longer start from scratch, and the LLM can continue conversations seamlessly across sessions. This KG is saved in form of raw triplets in the SQL database and in form of vector embeddings in a vector database along with relevant metadata used for filtering relevant set of triplets.

\subsubsection{Dynamic User Persona via KG}
In addition to storing conversation history, Memoria constructs a \textbf{dynamic user persona} by incrementally building a KG based on user interactions. This KG captures:

\begin{itemize}
    \item \textbf{Recurring topics} mentioned by the user;
    \item \textbf{User preferences} inferred from conversational patterns;
    \item \textbf{Named entities} identified during interactions; and,
    \item \textbf{Relationships} and connections between various user-stated facts.
\end{itemize}

The KG evolves as new messages are processed, enabling the system to emulate a growing, adaptive memory. This representation supports the generation of responses that are not only contextually accurate but also deeply personalized, based on the user's history and interaction patterns. Some of the key benefits of this KG are that the conversations feel customized to the end user, and meaningful responses are tailored rather than generic.

\subsubsection{Session Level Memory for Real Time Context}
To handle short-term recall within an ongoing session, Memoria includes a session summarization module. As the conversation progresses, this module dynamically generates summaries of the dialogue, which are then used to maintain continuity in real time.

This feature ensures that the LLM retains a coherent understanding of prior turns within the same session, crucial for multi-turn conversations where context evolves rapidly. By maintaining a live memory of the current session, Memoria mitigates context loss and allows the LLM to respond intelligently, even during extended dialogues.

\subsubsection{Seamless Retrieval for Context-Aware Responses}
Memoria's retrieval module enables the LLM to recall relevant past information when a user returns after a pause, whether the gap is a few minutes or several days. This retrieval mechanism fuses two memory sources:
(1) the \textbf{structured conversation history} from the database; and, (2) the \textbf{KG} constructed from prior interactions enhanced with weightage based on recency of conversations.

By effectively combining these, the system provides the LLM with a relevant distilled context window tailored to the current prompt. This enables the conversational assistant to recall previously discussed topics, user-specific preferences, and unfinished queries, allowing the conversation to resume naturally and productively. Both approaches, i.e., session-level summary and KG generation of user persona, help to improve the episodic and semantic memory capabilities of any LLM chat application.

\subsection{Holistic Personalization and Memory Retention}
Each of the above four modules in Memoria is designed to be independently functional yet synergistic when combined. Collectively, they form a holistic memory architecture that supports:

\begin{itemize}
    \item \textbf{Personalization:} By modeling individual users through KGs;
    \item \textbf{Retention:} By persistently storing conversation history;
    \item \textbf{Continuity:} By linking sessions via intelligent retrieval and summarization; and,
    \item \textbf{Efficiency:} By reducing redundancy and unnecessary user input.
\end{itemize}

Memoria functions as a plug-and-play memory extension for any LLM-based conversational assistant. It redefines user experience by transforming LLMs from forgetful responders into intelligent, adaptive agents capable of learning and evolving through ongoing interactions.

Memoria upgrades the LLM and user interaction paradigm from isolated exchanges to ongoing relationships and conversations that evolve, adapt, and remember.

\section{System Architecture}

\begin{figure*}[t]
    \centering
    \includegraphics[width=1\linewidth]{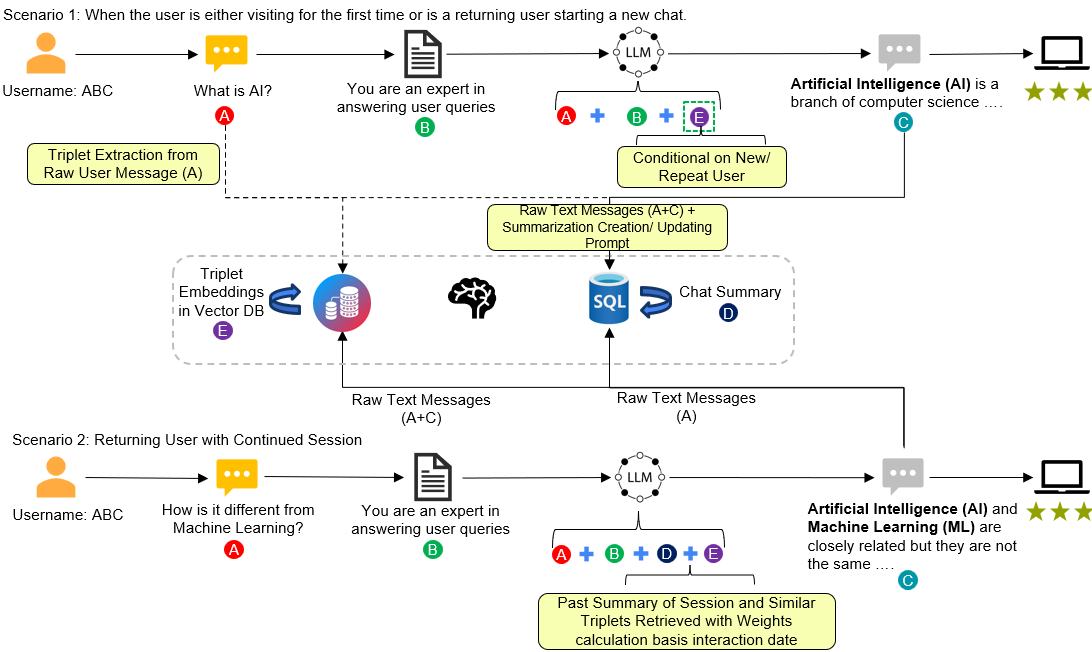}
    \caption{Memoria System Architecture}
    \label{fig:System Architecture}
\end{figure*}

The Memoria framework is designed to function as a modular memory enhancement layer for any LLM powered conversational system. Its architecture supports dynamic memory construction and retrieval based on the user’s session status distinguishing between new and repeat users, as well as new and ongoing sessions. Figure~\ref{fig:System Architecture} illustrates two operational scenarios that encapsulate Memoria's working components.

\subsection{Scenario 1: New User or New Session}
When a user interacts with the system for the first time, Memoria does not possess any prior memory, neither a session-level summary nor a users persona KG. As shown in the upper half of Figure~\ref{fig:System Architecture}, only the user message (A) and the system prompt (B) are passed to the LLM for response generation. No triplets (E) are involved at this point due to the absence of prior context. However, if the session is new but the user is a repeat visitor, the system augments the prompt with additional context derived from the user’s KG (E), alongside the user message and system prompt. Thus, inclusion of the KG in the prompt is conditional used only when prior user data exists from previous interactions.

Once a response is generated, the key memory update engine of Memoria is activated. The system performs the following steps:

\begin{itemize}
    \item Extracts knowledge triplets via LLM from the user message and stores them:
    \begin{itemize}
        \item In SQL for raw representation (subject, predicate, object).
        \item In vector database as vector embeddings, along with metadata (e.g., timestamp, original sentence, user name).
    \end{itemize}
    
    \item Saves the user message and assistant response in SQL as raw messages for full traceability.
    
    \item Triggers summary generation using both the user and assistant messages via an LLM call. The resulting summary is stored against the session ID in SQL.
\end{itemize}
This forms the foundational memory layer for any follow-up interactions within the same session.

\subsection{Scenario 2: Repeat User with Ongoing Session}

In the case of a repeat user continuing an ongoing session, Memoria retrieves both:

\begin{itemize}
    \item A session summary that encapsulates recent conversation turns.
    \item A set of user-specific triplets representing their KG.
    \item As shown in the lower half of Figure~\ref{fig:System Architecture}, the summary and the top-K triplets (retrieved via semantic similarity from vector DB) filtered for the user basis user name are retrieved. These triplets are further weighted in real time using an exponential decay function, giving higher priority to more recent triplets, ensuring that updated preferences or contradictions are resolved contextually. The session summary and weighted triplets are ingested into the prompt and further passed into the LLM for answering the user query, considering the additional context as well.
    \item Following response generation, the same update mechanism applies:
    \begin{itemize}
        \item New triplets from the user query are appended to SQL and embedded in the vector database.
        \item The raw messages and updated session summary are saved back to SQL.
    \end{itemize}
    \item This iterative loop continuously enriches both the long-term persona model and short-term conversation memory.
\end{itemize}

Memoria’s architecture ensures context-aware, consistent, and evolving dialogue with users while significantly reducing token overhead by avoiding full-history prompting.

\section{Working Mechanism}
Memoria's architecture is designed to manage memory across diverse user interaction patterns dynamically. This section outlines how the framework handles memory operations, including chat summarization, KG updating, and persistent storage across three distinct user scenarios in detail:

\begin{enumerate}
    \item New User with New Session;
    \item Repeat User with New Session; and,
    \item Repeat User with Continuing Session.
\end{enumerate}

Each scenario is processed through a consistent pipeline involving summary retrieval, KG access, application response capture, memory updates, and storage.

\subsection{New User with New Session}
When a user interacts with the system for the first time, the following events take place:

\begin{itemize}
    \item \textbf{Session Check:} Memoria verifies that no past summary exists for the session ID since it is a new session. Therefore, the summary retrieval module returns a null response.
    
    \item \textbf{User Check:} As this is a first-time user, no associated KG exists. The KG retrieval module confirms the absence of triplets or historical preferences.
    
    \item \textbf{Response Capture:} The user's query is passed to the developer’s application, which generates a response. Both the query and the corresponding reply are captured by Memoria.
    
    \item \textbf{Summary Generation:} Since no previous summary exists, a fresh session summary is constructed based on the user query and system response.
    
    \item \textbf{KG Extraction:} Relevant KG triplets are extracted solely from the user’s message. These triplets are embedded as dense vectors and stored in a vector database, along with metadata such as timestamps, user name and source message references etc.
    
    \item \textbf{Follow-up Retrieval:} For any subsequent queries within the same session, both the generated summary and updated KG are available. These can be retrieved using Memoria’s functions and used by the application developer to augment the system or user prompt, enhancing the coherence and personalization of follow-up answers.
\end{itemize}

\begin{figure}[H]
    \centering
    \includegraphics[width=1\linewidth, scale=1.2]{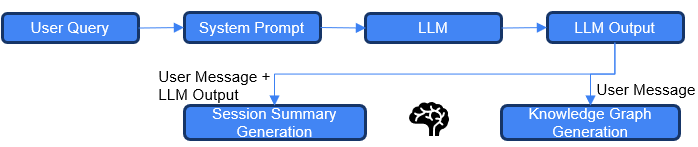}
    \caption{New User with New Session Flow}
    \label{fig:New User with New Session Flow}
\end{figure}

\subsection{Repeat User with New Session}
\begin{figure}[H]
    \centering
    \includegraphics[width=1\linewidth]{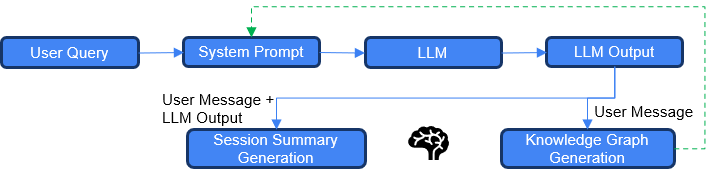}
    \caption{Repeat User with New Session Flow}
    \label{fig:Repeat User with New Session Flow}
\end{figure}

In this case, the user has interacted with the system before, but the current session is new:

\begin{itemize}
    \item \textbf{Session Summary:} As it is a new session, no prior summary is available, and the retrieval returns an empty response.

    \item \textbf{User Check:} Memoria identifies the user as a returning user and immediately retrieves the relevant user’s KG triplets based on similarity from previous interactions further assigned with weights (green dotted line in Figure \ref{fig:Repeat User with New Session Flow}).

    \item \textbf{Prompt Augmentation:} The application developer has access to these triplets from the first interaction of the session itself, enabling immediate personalization based on known preferences, interests, or past topics.

    \item \textbf{Memory Update:} As the conversation progresses, Memoria continuously updates both the session-level summary and the KG, allowing downstream interactions to benefit from enriched memory and contextual grounding.
\end{itemize}

\subsection{Repeat User with Continuing Session}
\begin{figure}[H]
    \centering
    \includegraphics[width=1\linewidth]{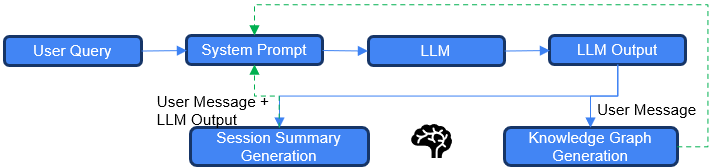}
    \caption{Repeat User with Repeat Session Flow}
    \label{fig:Repeat User with Repeat Session Flow}
\end{figure}

Here, the user is continuing within an already active session:

\begin{itemize}
    \item \textbf{Memory Availability:} Both session-level summaries and the user’s KG are readily available from the outset.

    \item \textbf{Retrieval and Prompt Construction:} The summary and relevant triplets, filtered and weighted (explained in the next section) are retrieved by the application and injected into the prompt to ensure deep context continuity (green dotted line in Figure \ref{fig:Repeat User with Repeat Session Flow}).

    \item \textbf{Enhanced Dialogue Flow:} With both long-term (KG) and short-term (session summary) context available, the LLM is primed to deliver highly coherent, context-aware, and personalized responses that evolve fluidly with the ongoing interaction.
\end{itemize}

\noindent This structured handling of different user scenarios ensures that Memoria offers scalable memory augmentation without compromising the user experience, regardless of whether the interaction is the user's first or hundredth.

\subsection{Memory Update Engine}
The core of Memoria’s memory system lies in its dual update mechanisms: one for maintaining session-level summaries and another for building a dynamic, personalized KG of the user. Both processes are tightly integrated with the LLM, ensuring that memory is not only persistently stored but also actively leveraged to support coherent and personalized conversations.

\subsubsection{Session Summary Update}

Upon receiving both the user query and the assistant's response, Memoria first evaluates whether a session-level summary already exists for the given session ID. If a prior summary is present, it is updated using the new pair of user and assistant messages through a summarization routine powered by the LLM. If no such summary exists, a new one is created by the LLM and stored in the database against the current session ID. This enables incremental construction of a coherent thread of dialogue over time. Retrieval of summaries is direct and deterministic, handled via session ID lookups through a dedicated library function that returns the current summary for any session.

\subsubsection{KG Update and Semantic Retrieval}

The KG component in Memoria represents a significant advancement in modeling persistent user-specific memory. Unlike the summary, the KG is constructed solely from the user’s input, excluding assistant responses, to ensure that the representation accurately reflects the user's intent, preferences, and identity. When a user message is received, the system checks whether an existing KG is available for that user based on the username. Regardless of the graph’s prior existence, structured triplets (subject, predicate, object) are extracted from the user’s query. If the graph already exists, the new triplets are intelligently connected to existing nodes to form an evolving semantic structure; otherwise, a new graph is instantiated if not KG exists for the user and are saved in SQL database.

Each extracted triplet is embedded into a vector space and stored in a vector database along with rich metadata. This includes the source user message, the timestamp of the conversation, and raw use message etc. At the time of retrieval, the incoming user query is converted into an embedding vector, and semantic similarity is used to retrieve the top $N$ most relevant triplets. These retrieved triplets provide contextual grounding for the LLM.

To dynamically prioritize KG triplets based on their recency, we apply an \textit{Exponential Weighted Average (EWA)} scheme. This method assigns higher weights to more recent triplets, while gradually decreasing the importance of older ones. Each triplet is assigned a weight based on how recently it was derived from the user's input messages. This approach prioritizes triplets that were mentioned more recently in the conversation, ensuring that the most up-to-date information is emphasized. It is particularly effective in resolving conflicts where the user's latest inputs contradict earlier facts, enabling accurate memory updates and maintaining consistency in the KG. Such decay-based weighting ensures that the model emphasizes the user's latest preferences and facts, which is critical in adapting to evolving contexts.

\subsection{Weight Calculation}

For each triplet $i$, we compute its raw weight $w_i$ using the following exponential decay function:

\begin{equation}
w_i = e^{-a \cdot x_i}
\end{equation}
where $a > 0$ is the \textbf{decay rate} that controls how quickly the weight decreases over time, and $x_i$ is the \textbf{number of minutes} between the current time and the creation date-time of triplet $i$. 
The values of $x$ representing the number of minutes between the current time and the creation time of a triplet are normalized to lie within the interval $[0, 1]$. This normalization step is crucial to ensure numerical stability and meaningful contribution of older triplets in the exponential weighting process.

Without normalization, large values of $x$ (which occur when the triplet originates from significantly older conversations) would lead to the exponential term $e^{-\alpha x}$ approaching zero. This would diminish the influence of otherwise important past interactions and effectively remove them from the model’s context.

By applying min-max normalization across all triplets:
\[
x_{\text{norm}} = \frac{x - x_{\min}}{x_{\max} - x_{\min}},
\]
we ensure that the exponential decay remains sensitive to relative recency, while preventing extreme suppression of older triplet weights. This allows the system to retain a soft memory of long-past interactions, especially when no newer, conflicting information exists.

A higher value of $a$ results in a steeper decay, meaning older triplets lose influence more rapidly. Conversely, smaller values of $a$ produce a gentler decay curve.

To ensure that the weights are comparable and sum to $1$, we normalize them across all retrieved triplets. The normalized weight $\tilde{w}_i$ is computed as:

\begin{equation}
\tilde{w}_i = \frac{w_i}{\sum_{j=1}^{N} w_j} = \frac{e^{-a \cdot x_i}}{\sum_{j=1}^{N} e^{-a \cdot x_j}},
\end{equation}
where $N$ is the total number of retrieved triplets. This normalization ensures that:
\begin{equation}
\sum_{i=1}^{N} \tilde{w}_i = 1
\end{equation}

These normalized weights are attached to their corresponding triplets when retrieved from the vector database. During prompt construction, they guide the LLM to prioritize triplets with higher weights, i.e., more recent and likely more relevant knowledge, thereby enhancing personalization and factual accuracy in the conversation. If conflicting triplets are retrieved, this weighting system enables the model to resolve discrepancies in favor of the most current knowledge, having higher weights.

\section{Experiments}
To evaluate Memoria against existing agentic memory systems, we compare its performance with A‑Mem \cite{xu2025mem}, a recently-proposed framework that organizes memory notes via principles inspired by the Zettelkasten method. A‑Mem automatically generates structured memory entries with contextual descriptions, tags, and embeddings, then dynamically links new memories to related historical entries and evolves its memory graph over time.

\subsection{Datasets}
For experimental evaluation, we utilized the LongMemEvals dataset\cite{wu2024longmemeval}, which is designed to benchmark memory-augmented language model systems in realistic business-oriented settings. The dataset contains long-form conversations paired with targeted questions, challenging the model's ability to retain and utilize contextual information along with ground truth answers for evaluation. On average, each conversation spans approximately 115,000 tokens, sufficiently large to test the boundaries of context retention while still falling within the token limits of current frontier LLMs\cite{rasmussen2025zep}.

LongMemEvals includes six distinct question types: (1) single-session-user, (2) single-session-assistant, (3) single-session-preference, (4) multi-session, (5) knowledge-update, and (6) temporal-reasoning. These categories represent various aspects of agentic memory and contextual reasoning. However, for this work, we restrict our analysis to the knowledge-update and single-session-user categories, as they align most closely with the core capabilities of Memoria. The dataset file used is longmemeval\_s available under LongMemEvals huggingface \cite{longmemeval2024}.

\subsection{Computational Details}
To align with the design goal of accessibility and open-source availability, Memoria has been implemented to operate entirely without reliance on external or proprietary databases. By default, the framework uses an on-premise SQLite3 database for storing raw user conversations, generated summaries, and associated metadata. For semantic representation of user traits and preferences, KG triplets are embedded and stored in a local instance of ChromaDB\footnote{https://docs.trychroma.com/docs/overview/introduction}, enabling vector-based retrieval with metadata support.

We employ OpenAI’s text-embedding-ada-002 model to generate embeddings for triplet storage and semantic similarity-based retrieval. The underlying language model used for generation and reasoning tasks is GPT-4.1-mini. During retrieval, we apply semantic top-$K$ matching with $K=20$, and assign temporal weights to retrieved triplets using an exponential decay function with a decay rate $\alpha$=0.02, prioritizing recent interactions.

The experiments were conducted without the need for high-performance or GPU-backed computational hardware due to the use of closed-source LLMs accessed via API endpoints, hence the system remains lightweight, cost-effective, and deployable in resource-constrained environments.

\subsection{Evaluation Setup}

We evaluated both Memoria and A-Mem on LongMemEvals’s \texttt{single-session-user} and \texttt{knowledge-update} subsets using GPT‑4.1‑mini as the LLM.

To provide a fair comparison, we executed two versions of A‑Mem:

\begin{enumerate}
  \item \textbf{Default A‑Mem}: Cloned the official repository \cite{agiresearch_amem} and ran experiments using its standard configuration, which employs the all-MiniLM-L6-v2 \cite{reimers-2020-allMiniLM-L6-v2} embedding model from SentenceTransformers.
  \item \textbf{Modified A‑Mem}: Updated the A‑Mem embedding component by replacing the default model with OpenAI’s text-embedding-ada-002, aligning with Memoria’s embedding strategy to have a direct comparison with Memoria, though no other changes were made.
\end{enumerate}

Memoria was tested using text-embedding-ada-002, with identical downstream configurations (GPT‑4.1‑mini, top‑20 retrieval, decay rate $\alpha=0.02$).

For a fair comparison, both Memoria and A‑Mem were configured to retrieve the same number of triplets/ messages (K = 20) in both the experiments that is, the default embedder (SentenceTransformers all-MiniLM-L6-v2) and modified version (OpenAI text-embedding-ada-002). A‑Mem does not implement a decay mechanism thus no weighting while Memoria applies its exponential decay-based weighting, a feature unique to our system.

\section{Results}

This section presents a comparative analysis of Memoria and A-Mem on the LongMemEval dataset, evaluating accuracy, token usage and latency performance.

\subsection{Accuracy}

Table \ref{tab:memoria_results} presents the performance comparison across four different approaches on two categories from the LongMemEvals dataset: \texttt{single-session-user} and \texttt{knowledge-update}. The evaluated methods include: (1) \textbf{Full Context}, where the LLM is prompted without memory augmentation, (2) \textbf{Memoria}, our proposed framework with summarization and KG retrieval, (3) \textbf{A‑Mem (ST)}, the original A‑Mem setup using SentenceTransformers, and (4) \textbf{A‑Mem (OpenAI)}, a modified A‑Mem configuration using OpenAI embeddings. All experiments were conducted using GPT-4.1-mini as the language model backend and evaluated with ground truth using LLM as a judge.

\begin{table}[h]
\centering
\resizebox{0.5\textwidth}{!}{
\begin{tabular}{|l|c|c|c|c|}
\hline
\textbf{Type} & \textbf{Full Conversation} & \textbf{A-Mem (ST)} & \textbf{A-Mem (OA)} & \textbf{Memoria}\\
\hline
Single-Session & 85.7\% &  78.5\% & 84.2\% & \textbf{87.1\%}\\
Knowledge-Update & 78.2\% & 76.2\% & 79.4\% & \textbf{80.8\%}\\
\hline
\end{tabular}}
\caption{Accuracy Comparison Across Memory Strategies}
\label{tab:memoria_results}
\end{table}

The results in Table~\ref{tab:memoria_results} highlight Memoria’s superior performance over both variants of A-Mem using SentenceTransformers (ST) and OpenAI (OA) embeddings across both Single-Session and Knowledge-Update scenarios. While full-context prompting performs reasonably well, it scales poorly due to latency and token cost. Memoria, by contrast, achieves the highest accuracy in both categories while significantly reducing prompt length by leveraging a weighted KG.

In our experiments, Memoria outperforms A-Mem by using recency-aware weighting for knowledge triplets. Unlike A-Mem’s unweighted retrieval, Memoria applies exponential decay to prioritize recent user inputs, resolve contradictions and emphasize updated information. This ensures coherent, up-to-date responses across sessions.

By combining structured long-term memory with adaptive prioritization, Memoria demonstrates that intelligent memory curation—rather than exhaustive recall—is a more effective and scalable strategy for memory-augmented LLMs.

\subsection{Latency Evaluation}
In addition to accuracy-based evaluations, we conducted a latency test to assess the end-to-end execution time for processing the complete evaluation dataset. This dataset comprises a total of $148$ data points, with $70$ samples under the \texttt{single-session-user} category and $78$ under the \texttt{knowledge-update} category.

We compare two approaches:
\begin{enumerate}
    \item \textbf{Baseline Approach (Full Context Prompting)}: In this setup, the entire historical conversation is appended to the prompt along with the current question. The LLM is then queried to generate an answer based on this complete conversational context.
    
    \item \textbf{Memoria-Augmented Approach}: In this setup, only the current question is provided to the system. The response time is calculated by aggregating the execution times of individual Memoria components, which include:
    \begin{itemize}
        \item Retrieval of relevant KG triplets,
        \item Computation of semantic similarity and filtering,
        \item Calculation of exponential decay-based weights for each triplet,
        \item Construction of the prompt using the weighted triplets,
        \item Final inference time from the LLM based on this constructed prompt.
    \end{itemize}
\end{enumerate}

\begin{table}[h]
\centering
\resizebox{0.5\textwidth}{!}{
\begin{tabular}{|c|c|c|c|}
\hline
Approach & Question Type & Inference Time & Avg. Token length \\
\hline
Full Context & single-session-user & 391 secs & 115K \\
Memoria & single-session-user & 260 secs & \textbf{398} \\
A-Mem (ST) & single-session-user & 290 secs & 958\\
A-Mem (OA) & single-session-user & \textbf{252 secs} & 934\\
\hline
Full Context & knowledge-update & 522 secs & 115K \\
Memoria & knowledge-update & \textbf{320 secs} & \textbf{400} \\
A-Mem (ST) & knowledge-update & 364 secs &  933\\
A-Mem (OA) & knowledge-update & 328 secs & 928\\

\hline
\end{tabular}}
\caption{Inference and Token Length Comparison Results between Full Context and Memoria.}
\label{tab:latency}
\end{table}

Table~\ref{tab:latency} presents the inference time and average prompt token length for different memory strategies across the two evaluated question types. Full-context prompting, while contextually rich, incurs the highest latency up to 522 seconds and processes over 115,000 tokens per query, resulting in substantial computational and monetary overhead. In contrast, Memoria reduces average token length to under 400 tokens by retrieving only a curated set of weighted KG triplets and session summaries, achieving up to 38.7\% reduction in latency compared to full context prompting.

Even though A-Mem variants offer good reductions in latency and prompt size compared to full context, their unweighted retrieval mechanism leads to less precise memory usage and higher average token lengths (900+ tokens) as it retrieves raw user messages for context. Memoria’s modular memory operations and recency-aware triplet weighting preserve contextual fidelity while remaining cost-efficient and scalable, particularly as session histories grow longer.

\section{Conclusion}

We propose an agentic memory framework, called Memoria, for personalized conversational AI. By enabling structured and persistent memory, Memoria allows LLMs to recall, reason, and act upon past interactions, leading to more personalized and coherent conversations. This addresses the statelessness problem commonly seen in traditional LLM deployments.

We will be releasing the library soon as an open-source Python package, built for seamless integration into existing infrastructures. 

Future work will expand Memoria's evaluation in key system dimensions, including memory footprint growth, retrieval latency under load, and KG quality metrics. We also aim to extend its use to broader agentic systems such as retrieval-based recommenders and productivity tools. By modularizing its components, Memoria can support advanced memory tasks like temporal reasoning, user preference tracking, and multisessioncoherence across diverse applications.

\bibliography{main}{}

@article{rasmussen2025zep,
  title={Zep: A Temporal Knowledge Graph Architecture for Agent Memory},
  author={Rasmussen, Preston and Paliychuk, Pavlo and Beauvais, Travis and Ryan, Jack and Chalef, Daniel},
  journal={arXiv preprint arXiv:2501.13956},
  year={2025}
}

@article{xu2025mem,
  title={A-mem: Agentic memory for llm agents},
  author={Xu, Wujiang and Mei, Kai and Gao, Hang and Tan, Juntao and Liang, Zujie and Zhang, Yongfeng},
  journal={arXiv preprint arXiv:2502.12110},
  year={2025}
}

@article{chang2024survey,
  title={A survey on evaluation of large language models},
  author={Chang, Yupeng and Wang, Xu and Wang, Jindong and Wu, Yuan and Yang, Linyi and Zhu, Kaijie and Chen, Hao and Yi, Xiaoyuan and Wang, Cunxiang and Wang, Yidong and others},
  journal={ACM transactions on intelligent systems and technology},
  volume={15},
  number={3},
  pages={1--45},
  year={2024},
  publisher={ACM New York, NY}
}

@article{wu2024longmemeval,
  title={LongMemEval: Benchmarking Chat Assistants on Long-Term Interactive Memory},
  author={Wu, Di and Wang, Hongwei and Yu, Wenhao and Zhang, Yuwei and Chang, Kai-Wei and Yu, Dong},
  journal={arXiv preprint arXiv:2410.10813},
  year={2024}
}

@article{luhmann1992communicating,
  title={Communicating with Slip Boxes. An Empirical Account},
  author={Luhmann, Niklas},
  journal={Two Essays by Niklas Luhmann},
  year={1992}
}

@article{baddeley2020working,
  title={Working memory},
  author={Baddeley, Alan},
  journal={Memory},
  pages={71--111},
  year={2020},
  publisher={Routledge}
}

@article{tulving1972episodic,
  title={Episodic and semantic memory},
  author={Tulving, Endel},
  journal={Organization of Memory},
  pages={381--403},
  year={1972},
  publisher={Academic Press}
}

@misc{llamaindex2024,
  author = {Jerry Liu and contributors},
  title = {LlamaIndex: Data Framework for LLM Applications},
  year = {2024},
  howpublished = {\url{https://docs.llamaindex.ai/en/stable/}},
  note = {Accessed: 2025-06-14}
}

@article{zhang2024survey,
  title={A survey on the memory mechanism of large language model based agents},
  author={Zhang, Zeyu and Bo, Xiaohe and Ma, Chen and Li, Rui and Chen, Xu and Dai, Quanyu and Zhu, Jieming and Dong, Zhenhua and Wen, Ji-Rong},
  journal={arXiv preprint arXiv:2404.13501},
  year={2024}
}

@misc{longmemeval2024,
  author       = {Xiaowu and Pengcheng Yin and Graham Neubig},
  title        = {LongMemEval: A Benchmark for Long-Term Memory in Language Models},
  howpublished = {\url{https://huggingface.co/datasets/xiaowu0162/longmemeval}},
  year         = {2024},
  note         = {Accessed: 2025-06-25}
}

@misc{reimers-2020-allMiniLM-L6-v2,
  author       = {Nils Reimers and Iryna Gurevych},
  title        = {Sentence-Transformers — all-MiniLM-L6-v2},
  howpublished = {\url{https://huggingface.co/sentence-transformers/all-MiniLM-L6-v2}},
  note         = {Accessed: 2025-07-10},
  year         = {2020},
  organization = {Hugging Face},
}

@misc{agiresearch_amem,
  title        = {A​-MEM: Agentic Memory for LLM Agents},
  author       = {Xu, Wujiang and Liang, Zujie and Mei, Kai and Gao, Hang and Tan, Juntao and Zhang, Yongfeng},
  howpublished = {\url{https://github.com/agiresearch/A-mem}},
  note         = {GitHub repository; last accessed 10 July 2025},
  year         = {2025}
}

@misc{langgraph2025,
  title        = {LangGraph},
  howpublished = {\url{https://www.langchain.com/langgraph}},
  note         = {LangChain, accessed July 10, 2025},
  year         = {2025},
  organization = {LangChain}
}
\bibliographystyle{unsrt}
\end{document}